\documentclass[10pt,twocolumn,letterpaper]{article}

\usepackage{iccv}
\usepackage{times}
\usepackage{epsfig}
\usepackage{graphicx}
\usepackage{amsmath}
\usepackage{amssymb}

\usepackage[utf8]{inputenc}
\usepackage[T1]{fontenc}
\usepackage{textcomp}

\usepackage{times}
\usepackage{gensymb}
\usepackage{graphicx}
\usepackage{amsmath}
\usepackage{amssymb}
\usepackage{comment}
\usepackage{capt-of}
\usepackage{microtype}
\usepackage{xcolor}

\usepackage{algorithm}
\usepackage{algpseudocode}
\usepackage{mathrsfs}
\usepackage{multirow}
\usepackage{colortbl}
\usepackage[subtle]{savetrees}
\usepackage{tabu}

\usepackage{array}
\newcolumntype{L}[1]{>{\raggedright\let\newline\\\arraybackslash\hspace{0pt}}m{#1}}
\newcolumntype{C}[1]{>{\centering\let\newline\\\arraybackslash\hspace{0pt}}m{#1}}
\newcolumntype{R}[1]{>{\raggedleft\let\newline\\\arraybackslash\hspace{0pt}}m{#1}}

\definecolor{skyblue}{rgb}{0.2,0.6,0.9}

\newcommand{\param}{\theta}
\newcommand{\params}{\boldsymbol{\param}}

\DeclareMathOperator*{\argmin}{arg\,min}
\newcommand{\norm}[1]{\left\lVert#1\right\rVert}

\usepackage[pagebackref=true,breaklinks=true,letterpaper=true,colorlinks,bookmarks=false]{hyperref}

\iccvfinalcopy 

\ificcvfinal\fi
\begin{document}

\title{EgoFace: Egocentric Face Performance Capture and Videorealistic Reenactment}

\author{Mohamed Elgharib~~~~~Mallikarjun BR~~~~~Ayush Tewari \\ 
Hyeongwoo Kim~~~~~Wentao Liu~~~~~Hans-Peter Seidel~~~~~Christian Theobalt \vspace{0.17cm} \\ 
        Max Planck Institute for Informatics
}

\twocolumn[{
	\renewcommand\twocolumn[1][]{#1}
	\maketitle
	\vspace{-1em}
	{\centering
		\includegraphics[width=1\linewidth]{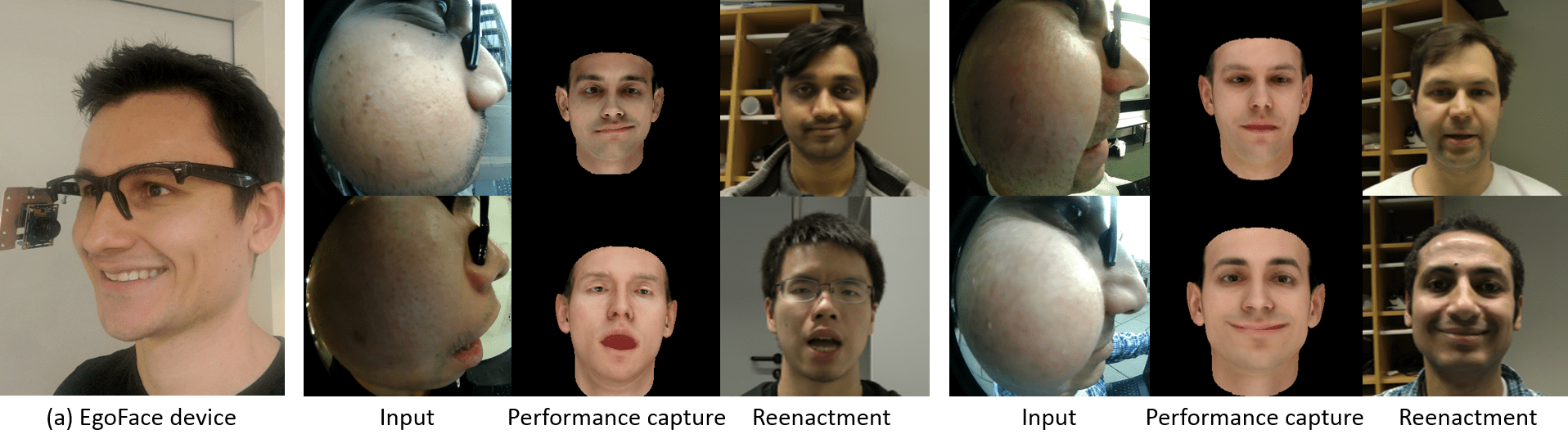}
		\captionof{figure}{%
			EgoFace uses a radically lightweight single RGB camera attached to an eye glass frame. It captures face performance and produces a front-view videorealistic reenactment directly from a single highly distorted one-sided input image. EgoFace handles a wide variety of expressions under varying illuminations, backgrounds and movements.			
        }
		\label{fig:teaser}
	}
	\vspace{1em}
}]

\maketitle

\begin{abstract}
   Face performance capture and reenactment techniques use multiple cameras and sensors, positioned at a distance from the face or mounted on heavy wearable devices. This limits their applications in mobile and outdoor environments. We present EgoFace, a radically new lightweight setup for face performance capture and front-view videorealistic reenactment using a single egocentric RGB camera. 
   Our lightweight setup allows operations in uncontrolled environments, and lends itself to telepresence applications such as video-conferencing from dynamic environments. 
   The input image is projected into a low dimensional latent space of the facial expression parameters. Through careful adversarial training of the parameter-space synthetic rendering, a videorealistic animation is produced. Our problem is challenging as the human visual system is sensitive to the smallest face irregularities that could occur in the final results. This sensitivity is even stronger for video results.  
   Our solution is trained in a pre-processing stage, through a supervised manner without manual annotations. EgoFace captures a wide variety of facial expressions, including mouth movements and asymmetrical expressions. It works under varying illuminations, background, movements, handles people from different ethnicities and can operate in real time. 
\end{abstract}


\section{Introduction}

Face performance capture algorithms reconstruct the dense dynamic face geometry and appearance from video footage.
They are crucial in many application areas, for instance, for the creation of digital actors in movies and VFX, as well as for 
the animation of avatars in virtual augmented reality.
Highest quality capture results are obtained with methods using dense multi-camera systems \cite{Beeler11}. However, such setups are expensive, immobile, and hard to use in ubiquitous applications \cite{klehm15star}. 
 Therefore, researchers developed face performance capture methods using lightweight sensors, such as a stereo camera \cite{VWBS12}, a depth camera \cite{Thies15,thies18headon}, or a single color camera \cite{ThiesZSTN2016a}. While results of these methods are less detailed than with the aforementioned multi-camera approaches, they are often real-time capable and much simpler to use in general everyday settings. 

Performance-driven face animation algorithms map captured face performances onto rendered digital doubles \cite{Alexander10} 
or avatar \cite{cao2016real,hu2017avatar}. 
A special case of such a mapping is face reenactment, which maps a captured face performance videorealistically onto a video of a person \cite{thies18facevr,ThiesZSTN2016a,thies18headon,kim18,Wu2018}. To simplify face performance capture for animation in movies and games, commercial helmet rigs with cameras mounted in front of the face were developed. For applications in VR, headsets were equipped with external cameras and strain gauges to enable performance-driven animation \cite{li15facial}, or with external and internal cameras to enable video reenactment-based headset removal for VR teleconferencing \cite{thies18facevr}. 

However, even recent lightweight face performance capture setups are obstructive and prevent many ubiquitous applications of face performance capture and performance-driven face animation that have great potential. Stereo helmet rigs are cumbersome, obstruct the field of view and are unusable in everyday surroundings. Even permanently holding up a single video or mobile phone camera in front of the face can be prohibitive for face performance capture for visual content creation, as it obstructs the face, and it is practically infeasible in many day-to-day situations. 

We therefore present EgoFace, a new approach for face performance capture and real-time performance-driven (videorealistic) face reenactment that uses a radically simpler, lightweight and unobstructive capture setup. EgoFace uses a single fisheye camera closely placed near a human face on a regular eye glass frame to the point it is not obstructing the user view. 
We present a new learning-based approach to estimate the expression parameters of a parametric face model 
from this highly challenging, highly distorted and only partial view of the face. Our approach captures a wide variety of mouth movements and asymmetric face expressions despite a one-sided lateral view of the face. An adversarially trained neural network then translates renderings of the captured face model into a videorealistic animation of the face from a frontal perspective (see Fig.~\ref{fig:teaser}). Facial performance capture and reenactment from our oblique egocentric view is challenging as current facial processing techniques are not designed to handle it. In addition, the human visual system is sensitive to the slightest irregularities on human faces that could occur in the results. This sensitivity is emphasized when the results are video. 

We make the following contributions:
\begin{enumerate}
    \item A radically lightweight setup and algorithm for egocentric face performance capture, enabling novel mobile facial animation applications 
    \item A method for videorealistic frontal view facial reenactment driven by the egocentric capture result 
    \item A dataset for egocentric facial performance capture and videorealistic reenactment
    \item An efficient implementation allowing operation in real-time for the full pipeline.
\end{enumerate}

\section{Related Work}

\begin{figure*}
\centering
	\includegraphics[width=0.9\linewidth]{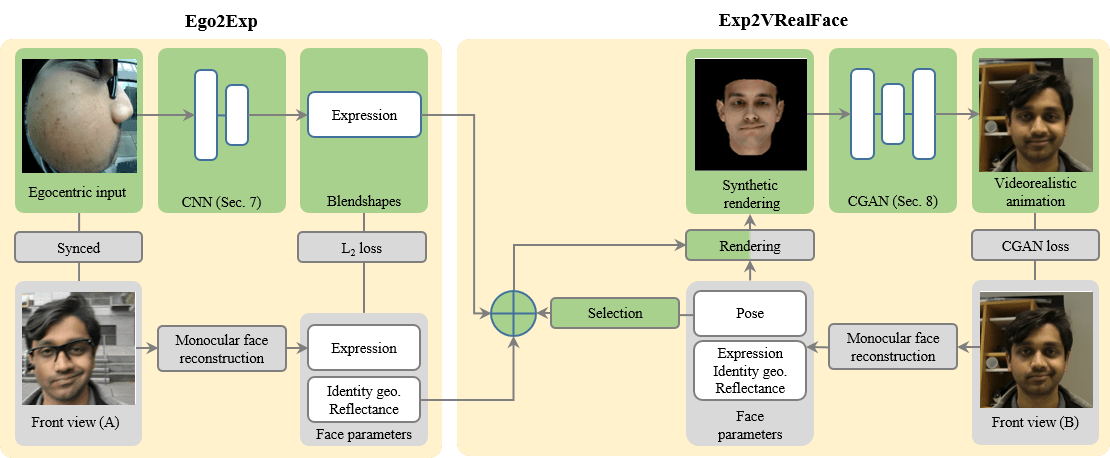}
	\caption{\label{fig:pipeline}
		EgoFace takes a single egocentric view as input, estimates face expressions through a CNN (Ego2Exp) and produces a videorealistic reenactment from the front view by utilizing a CGAN (Exp2VRealFace). Green blocks are used during testing, gray blocks are for training only. Ego2Exp is trained on paired egocentric and front-view (A) images while Exp2VRealFace is trained on paired front-views (B) and their synthetic renderings. Both networks are trained per person. During testing the user has control over the target pose through a selection block.
	}
\end{figure*}

Face capture and reenactment techniques can be classified into two categories; using wearable devices such as VR headsets \cite{thies18facevr,li15facial,Olszewski2016HFS,Cha2018TFM,Lombardi18DAM,Asano18FGA} and using non-wearable devices \cite{thies18headon,kim18,ThiesZSTN2016a,Thies15,SuwajSK2015,Vlasic05}. Wearable device techniques deal with partial facial occlusions caused by the head mounted displays. On the other hand, techniques for non-wearable devices require full facial coverage at once. We focus on methods using consumer level setups and sensors such as RGB, IR and depth cameras. Techniques utilizing production-level setups are out the scope of this work. Please refer to \cite{klehm15star} for more detail. 

\subsection{Non-Wearable Devices}

Thies et al. \cite{Thies15} presented the first real-time expression transfer approach using an RGB-D camera. For each frame, a parametric model for identity, expression, skin reflectance and lighting was fitted to the input. Expression is transferred by computing the difference between the source and target expressions in parameter space, and modifying the target parameters to match the source expressions. Later, Face2Face \cite{ThiesZSTN2016a} introduced expression transfer in real-time from a single RGB camera. Facial expressions of both source and target are tracked using a dense photometric consistency loss. Reenactment is achieved by fast and efficient deformation transfer between source and target. A learning-based approach is used to estimate the mouth interior for the retargeted expression. 

While the aforementioned approaches produce compelling results, they only control facial expressions. Thies et al. presented HeadOn \cite{thies18headon}, a technique allowing control of the 3D pose and gaze, aswell as the expressions. Here, a personalized geometry proxy that embeds a parametric head, eye, and kinematic torso model is used. While HeadOn generates very interesting results, it requires a depth camera. Kim et al. \cite{kim18} presented Deep Video Portraits, the first approach for controlling the 3D pose, facial expressions and gaze from a single RGB camera. Their approach is based on neural networks where they show generative adversarial networks can produce visually pleasing facial animations, inspired by the success of GANs \cite{Isola17}. 
A number of approaches produce video reenactment even given a single target image \cite{Nagano2018,Geng2018,Wiles18}. 
While \cite{Geng2018,Wiles18} can only control the face expressions, \cite{Nagano2018} can reenact both pose and expressions of the examined person.  
All of these approaches assume a close to frontal face view, while we deal with a highly distorted and occluded view of the face. 

\subsection{Wearable Devices}

A number of facial reenactment techniques for wearable device have recently been proposed \cite{thies18facevr,li15facial,Olszewski2016HFS,Cha2018TFM,Lombardi18DAM,Asano18FGA}, most of which are designed for VR headsets \cite{thies18facevr,li15facial,Olszewski2016HFS,Lombardi18DAM}. Such devices occlude the upper half of the face, and hence these techniques attempt to capture the full face through multiple sensors. Many approaches utilize an IR camera placed inside the headset to capture the eye movement \cite{thies18facevr,li15facial,Olszewski2016HFS}. Li et al. \cite{li15facial} placed eight ultra-thin strain gauges (flexible metal foil sensors) on the foam liner of the headset. This measures surface strain caused by the upper facial expressions. An RGB-D camera is also used to capture the lower facial expressions. Olszewski et al. \cite{Olszewski2016HFS} extended Li et al. \cite{li15facial} by exploiting the audio signal. Here, same utterance among different subjects is mapped to the same animation parameters and a single RGB camera is used to track the mouth movement. 

Thies et al. \cite{thies18facevr} presented FaceVR, a solution utilizing the optimization based infrastructure of \cite{ThiesZSTN2016a,Thies15}. RGB-D cameras are used to build a personalized stereo avatar by solving for a face parametric model. During tracking, expressions, lighting and rigid pose are optimized. Eye-gaze tracking is performed through random ferns while data-driven solution is used for mouth retrieval, similar to Face2Face \cite{ThiesZSTN2016a}. Recently, Lombardi et al. \cite{Lombardi18DAM} presented a deep appearance model for human face rendering, and showed how to control it through a VR headset. The model is estimated using multi-view light stage data. The headset is equipped with 3 cameras, 1 for the mouth and 2 for the eyes. A learning-based approach controls the face model, trained with synthetic facial expression images. Asano et al. \cite{Asano18FGA} proposed an eye-wear for face capture. It uses 20 photo-reflective sensors, 16 placed on the glass frame and 4 in the lower face region. Mo-cap markers are placed on the face, and a relation is learned between their positions and the measurements of the photo-reflective sensors. At testing markers are removed and the learned model deforms a synthetic avatar. Cha et al. \cite{Cha2018TFM} used two head-mounted cameras for face performance capture. Although interesting results are produced, the final output is a synthetic avatar.

\section{Overview}

EgoFace is a lightweight approach for face performance capture and videorealistic face reenactment using only a single RGB camera mounted to an eye glass frame (see Fig.~\ref{fig:teaser}). EgoFace consists of two stages:
1) extracting facial expression parameters (see Sec.~\ref{sec:Ego2Exp}) and 2) translating the resulting expression parameters to a videorealistic full-face from the front perspective (see Sec.~\ref{sec:Exp2VRealFace}). The first stage, Ego2Exp, extracts the facial expressions through a deep encoder (Fig.~\ref{fig:pipeline}, left). The second step, Exp2VRealFace, utilizes adversarial training to generate a videorealistic frontal-view of the user retargeted on a different environment (Fig.~\ref{fig:pipeline}, right). Each stage of our system is trained separately in a supervised manner without manual annotations (Sec.~\ref{sec:DataCollection}). EgoFace is person specific, with each network trained per person. We first discuss the parametric face model used through out the paper and how it can reconstruct a moving face. We then discuss our camera setup, followed by data collection. Ego2Exp and Exp2VRealFace are presented, followed by results and conclusion.

\section{Face Model and Reconstruction}
\label{sec:Face-Model}

We parameterize faces images using $m = \text{353}$ semantic parameters:
%
\begin{equation}\small
  \params = \left( \mathbf{R}, \mathbf{T}, \mathbf{\alpha}, \boldsymbol{\beta}, \boldsymbol\delta, \boldsymbol\gamma \right) \in \mathbb{R}^{m} \text{.}
\label{eq:fmodel}
\end{equation}
Here, global pose is represented using $\mathbf{R}$, which specifies the head rotation using Euler angles and  $\mathbf{T}$, which represents the head translation. The face geometry is represented using $\boldsymbol{\alpha}$, the identity geometry parameters and $\boldsymbol\delta$ the expression coefficients. $\boldsymbol{\beta}$ and $\boldsymbol{\gamma}$ define the reflectance of the face and the incident illumination respectively. 
We use $|\mathbf{R}|=3$,  $|\mathbf{T}|=3$, $|\boldsymbol{\alpha}|=128$, $|\boldsymbol{\beta}|=128$ and $|\boldsymbol\delta|=64$. 
The geometry of the face is described using stacked per-vertex positions of a template mesh, $\mathbf{v} \!\in\! \mathbb{R}^{3N}$ with $N=60\text{K}$ vertices.
\begin{equation}
\mathbf{v}(\boldsymbol\alpha, \boldsymbol\delta) = \mathbf{a}_\mathrm{geo} + \sum_{k=1}^{|\boldsymbol{\alpha}|}{\boldsymbol\alpha_k \mathbf{b}^\mathrm{geo}_k} + \sum_{k=1}^{|\boldsymbol{\delta}|}{\boldsymbol\delta_k \mathbf{b}^\mathrm{exp}_k} \text{.}
\end{equation}
Similarly, skin reflectance is defined using  stacked per-vertex reflectance, $\mathbf{r} \!\in\! \mathbb{R}^{3N}$ as:
\begin{equation}
\mathbf{r}(\boldsymbol\beta) = \mathbf{a}_\mathrm{ref} + \sum_{k=1}^{|\boldsymbol{\beta}|}{\boldsymbol\beta_k \mathbf{b}^\mathrm{ref}_k} \text{.}
\end{equation}
$\mathbf{a}_\mathrm{geo} \!\in\! \mathbb{R}^{3N}$ and $\mathbf{a}_\mathrm{ref} \!\in\! \mathbb{R}^{3N}$ are the average facial geometry and the skin reflectance, respectively. 
$\mathbf{b}^\mathrm{geo}_k$ and $\mathbf{b}^\mathrm{ref}_k \!\in\! \mathbb{R}^{3N}$ are the geometry and reflectance deformation models, learned using 200 3D face scans \cite{Blanz1999}.
 The expression model, $\mathbf{b}^\mathrm{exp}_k$ is a combination of blendshapes from the Digital Emily model \cite{Alexander10} and FaceWarehouse \cite{Cao2014b}. We use PCA to compress the original over-complete blendshapes to a subspace of $|\boldsymbol{\delta}|=64$ dimensions. We assume the face to be Lambertian and the illumination to be distant. The illumination is modeled using second-order spherical harmonics (SH) \cite{Ramamoorthi2001b}.
The diffused color of the $i$-th vertex, $\mathbf{c}_i$ is then computed through $\mathbf{c}_i(\mathbf{r}_i, \mathbf{n}_i, \boldsymbol\gamma) = \mathbf{r}_i \cdot \sum_{b=1}^{B^2} \boldsymbol\gamma_b Y_b(\mathbf{n}_i)$.
Here, $B=3$ represents the number of SH bands, $\boldsymbol\gamma_b \!\in\! \mathbb{R}^3$ are the SH coefficients of the basis functions $Y_b$, and $\mathbf{r}_i$ and $\mathbf{n}_i$ are the reflectance and normal vector of the $i$-th vertex.

\textbf{Monocular Face Reconstruction} We estimate the face parameters of a face in a front-view video by maximizing the photo-consistency between a synthetic rendering of the model and the input, akin to \cite{GarriZCVVPT2016,ThiesZSTN2016a}. The reconstruction energy contain terms for per vertex photo-consistency, landmark points alignment and statistical prior. 
This allows reconstruction of the identity (geometry and skin reflectance), facial expression, and scene illumination in a video. 
The sparse alignment term is defined on 66 automatically detected facial landmarks \cite{SaragLC2011}.
We use a perspective camera to project the 3D mesh onto the image plane. 
The identity geometry and reflectance parameters are estimated for the first frame and fixed for the rest of the sequence. All other face parameters are estimated for each frame.
For more details on the energy formulation and optimization, please refer to \cite{GarriZCVVPT2016,ThiesZSTN2016a}.

\section{Camera Setup}
\label{sec:camera-setup}

We designed a lightweight egocentric camera setup to enable face capture and videorealistic reenactment in mobile and uncontrolled environments. We use a single fish-eye commodity camera mounted on the frame of an eye glass. The camera is placed closely to the face to the point it does not obstruct the user view nor its movement (see Fig.~\ref{fig:teaser}-a). The camera weighs 9.1 grams, with maximum dimensions of $ 3.7 \!\times\! 3.7 \!\times\! 1.6$ cm, and can be easily obtained from consumer shops. The camera has a 170 degree field of view, capturing the face from the lower chin up to the upper part of the eye. The camera-face relative position is fixed, irrespective of the head orientation. This removes the need of estimating the head pose parameters during performance capture (Sec.~\ref{sec:Ego2Exp}). 

\section{Data Collection}
\label{sec:DataCollection}

\begin{figure*}
	\includegraphics[width=\linewidth]{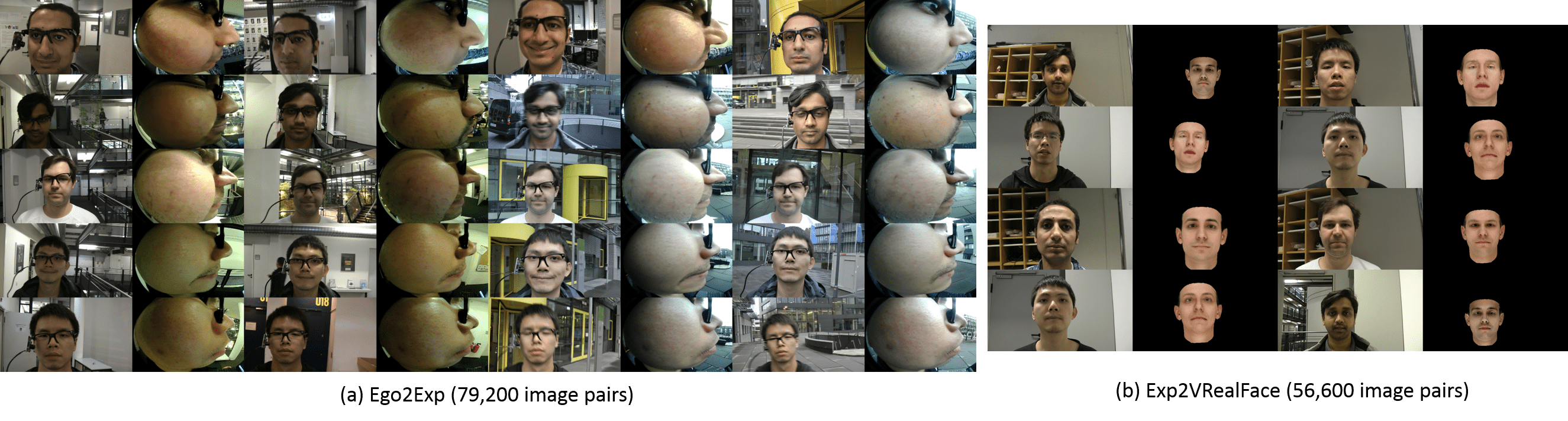}
	\caption{%
		Samples from our datasets. For both sets we provide the parameters of the face model (Eq.~\ref{eq:fmodel}). Ego2Exp contains synchronized egocentric and front view frames, shot in mobile indoor and outdoor environment, under several illuminations and backgrounds. Exp2VRealFace contain front view recordings and the corresponding albedo rendering in static environments. 
	}
	\label{fig:Data}
\end{figure*}

Our system has two camera collection setups, one for each stage of our pipeline (see Fig.~\ref{fig:Data}). For the first stage, Ego2Exp, we collect data simultaneously from the egocentric camera and from a front-view camera (see Fig.~\ref{fig:Data}-a). The user wears the egocentric headset, looks at the front-camera, walks and performs several facial expressions. We track the face from the front view camera using automated monocular face reconstruction (Sec.~\ref{sec:Face-Model}). 
Identity geometry and reflectance are estimated from one frame, and fixed in all remaining data of the same person. 
The resulting expression parameters are used as a supervising signal for training Ego2Exp with the egocentric view as input. We use a transient event to ensure both egocentric and front-view camera data are synchronized. A mobile phone screen observed from both cameras plays a video sequence of mostly black frames, with white frames every 10 seconds. During frame extraction, we make sure both cameras see the start of the white frames within 1 frame accuracy. We also use the same transient event at the end of the recording for verification purposes. Fig.~\ref{fig:Data} shows samples from our Ego2Exp training set. It contains 79,200 frames, at 25 fps and an original resolution of $1920 \!\times\! 1080$. It is shot under several indoor and outdoor environments, and contains varying illuminations and backgrounds. 

For the second stage, Exp2VRealFace, we collect data using only a front-view camera. The camera is positioned on a fixed tripod and the user sits on a chair, looks directly at the camera without wearing the egocentric headset (see Fig.~\ref{fig:Data}-b). The user performs several expressions and the face gets tracked using the method in Sec.~\ref{sec:Face-Model}. We fix the identity parameters for each subject. The corresponding synthetic albedo rendering is produced and used as input to Exp2VRealFace to learn a videorealistic face. We do not render the scene illumination and instead rely on the synthetic-to-real translation network to learn it (Sec.~\ref{sec:Exp2VRealFace}). The dataset contain 56,600 frames, at 25 fps and original resolution of $1920 \!\times\! 1080$, shot in static environments, with different backgrounds and illuminations.

\section{Deep Egocentric Expression Estimation}
\label{sec:Ego2Exp}

Given training data $\{\mathbf{I}_i, \boldsymbol{\delta}_i\}_{i=1}^N$ consisting of $N$ egocentric images $\mathbf{I}$ and the corresponding ground-truth expression $\boldsymbol{\delta}$ (Sec.~\ref{sec:DataCollection}), we train a deep encoder to infer the underlying expressions from the input $\mathbf{I}$. We examined different networks based on the popular 
AlexNet \cite{KrizhSH2012}, ResNet50 \cite{HeZRS2016} and VGG \cite{Simonyan14c} architectures. All networks were pre-trained on ImageNet classification \cite{KrizhSH2012}. Experiments show that VGG produces the best visual quality and hence we used it to display most of the results. We examined AlexNet and ResNet50 for achieving a real-time performance for the entire EgoFace pipeline (see Sec.~\ref{sec:results}, Real-time implementation). 
We resize the last fully-connected layer of VGG to match the dimensionality of 64 for the expressions blendshapes. We resize the input images to $224 \!\times\! 224$ to match the VGG input size. We reduce the impact of the background through an approximate binary face mask. We manually mask one frame for each sequence, which is sufficient due to the fixed camera position. We minimize for the mean squared difference between the ground-truth and estimated expressions. We train our network using the TensorFlow \cite{Abadi15}. We use ADAM solver \cite{Kingma2015AdamAM} with a batch size of $32$, dropout of $0.5$ and train for 50 epochs.

\section{Expressions to Videorealistic Full Face}
\label{sec:Exp2VRealFace}

The second stage of our system is Exp2VRealFace, a method for generating videorealistic front view face animation driven by the egocentric expressions. We first render a synthetic face of the user from the front perspective while maintaining the expressions estimated by Ego2Exp.
Since EgoFace is person-specific, the identity geometry and reflectance parameters are fixed per user, and therefore used in our front view rendering. We select a target sequence from the Exp2VRealFace dataset, and copy its head rotation and translation parameters (Fig.~\ref{fig:pipeline}, Selection). We either randomly choose one pose or iteratively loop over a small set of poses to allow small head sway. Given the full face parameters, we render the reflectance assuming a full perspective camera model. We do not account for illumination and instead rely on the Exp2VRealFace network to add it. 

The Exp2VRealFace network is inspired by the recent success of Kim et al. \cite{kim18} in producing videorealistic head portraits. 
A conditional generative adversarial network consisting of a generator network $\mathbf{G}$ and a discriminator $\mathbf{D}$ is used.
The generator network takes the synthetic rendering $\mathbf{X}$ as input and produces a videorealistic version $\mathbf{G}(\mathbf{X})$. 
We train this network in a supervised manner with paired samples (see Sec.~\ref{sec:DataCollection}). It is trained in a way to optimize 
\begin{equation}
\label{eq:Exp2VRealFace}
\argmin_{\mathbf{G}}~\max_{\mathbf{D}}~{E_{\textrm{A}}(\mathbf{G}, \mathbf{D}) + \lambda E_{\ell_1}(\mathbf{G})} \text{.}
\end{equation}
Eq.~(\ref{eq:Exp2VRealFace}) contains an adversarial loss $E_{\textrm{A}}(\mathbf{G}, \mathbf{D})$ and an $\ell_1$-norm reproduction loss $E_{\ell_1}(\mathbf{G})$.
$\lambda$ weighs the importance of each term and it is fixed to 10 in our experiments. The $\ell_1$-norm loss enforces the output image $\mathbf{G}(\mathbf{X})$ to resemble the ground-truth $\mathbf{Y}$ through
\begin{equation}
E_{\ell_1}(\mathbf{G}) =  \mathbb{E}_{\mathbf{X},\mathbf{Y}}\big[ \norm{ \mathbf{Y} - \mathbf{G}(\mathbf{X}) }_1 \big] \text{.}
\end{equation}
The adversarial loss is defined as:
\begin{multline}
E_{\textrm{A}}(\mathbf{G},\mathbf{D}) =
\mathbb{E}_{\mathbf{X},\mathbf{Y}} \big[ \!\log \mathbf{D}(\mathbf{X},\mathbf{Y}) \big] + \\
\mathbb{E}_{\mathbf{X}} \big[ \!\log\big(1 - \mathbf{D}(\mathbf{X},\mathbf{G}(\mathbf{X}))\big) \big] \text{.}
\end{multline}
The input to the discriminator $\mathbf{D}$ is the synthetic rendering $\mathbf{X}$, and either the predicted output image $\mathbf{G}(\mathbf{X})$ or the ground-truth image $\mathbf{Y}$.
The generator $\mathbf{G}$ takes as input synthetic renderings $\mathbf{X}$ to translate them into videorealistic outputs $\mathbf{G}(\mathbf{X})$ (see Fig.~\ref{fig:pipeline}, right). 
The Exp2VRealFace network is optimized by the principle of game theory~\cite{goodfellow2014generative,Isola17}, in which the generator minimizes the adversarial loss to provide outputs at a high level of videorealism, whilst the discriminator maximizes the classification accuracy of the generated outputs.

The architecture of encoder-decoder is described in Table.~\ref{table:arch-cgan}. Please refer to \cite{kim18} for full architecture. Skip connections are used to capture fine-details. To ensure temporal coherency, the network uses a moving temporal window by taking multiple frames as input and producing one frame at a time.
We also experimented with a more efficient architecture to run
the full EgoFace pipeline in real-time (see Sec.~\ref{sec:results}, Real-time implementation). We train our network using TensorFlow \cite{Abadi15}. We use ADAM solver \cite{Kingma2015AdamAM} with a batch size of $12$, learning rate of $0.0002$, and train for 200 epochs. 

\begin{table}[]
	\centering
	\small
	\resizebox{0.48\textwidth}{!}{%
		\begin{tabu}{c|[1pt]c|c|[1pt]c|c}
			         & \multicolumn{2}{c|[1pt]}{Full}   & \multicolumn{2}{c}{Optimized}                       \\ \tabucline[1pt]{-}
			Layer    & Output size      & Channels & Output size           & Channels                         \\ \tabucline[1pt]{-}
			Encoder1 & 128 $\times$ 128 & 64       & \multicolumn{2}{c}{\cellcolor{gray!50}}                  \\ \hline
			Encoder2 & 64 $\times$ 64   & 128      & 64 $\times$ 64        & 128                              \\ \hline
			Encoder3 & 32 $\times$ 32   & 256      & 32 $\times$ 32        & 256                              \\ \hline
			Encoder4 & 16 $\times$ 16   & 512      & 16 $\times$ 16        & 512                              \\ \hline
			Encoder5 & 8 $\times$ 8     & 512      & 8 $\times$ 8          & 512                              \\ \hline
			Encoder6 & 4 $\times$ 4     & 512      & \multicolumn{2}{c}{\multirow{4}{*}{\cellcolor{gray!50}}} \\ \cline{1-3}
			Encoder7 & 2 $\times$ 2     & 512      & \multicolumn{2}{c}{\cellcolor{gray!50}}                  \\ \cline{1-3}
			Decoder7 & 2 $\times$ 2     & 512      & \multicolumn{2}{c}{\cellcolor{gray!50}}                  \\ \cline{1-3}
			Decoder6 & 4 $\times$ 4     & 512      & \multicolumn{2}{c}{\cellcolor{gray!50}}                  \\ \hline
			Decoder5 & 8 $\times$ 8     & 512      & 8 $\times$ 8          & 512                              \\ \hline
			Decoder4 & 16 $\times$ 16   & 512      & 16 $\times$ 16        & 512                              \\ \hline
			Decoder3 & 32 $\times$ 32   & 256      & 32 $\times$ 32        & 256                              \\ \hline
			Decoder2 & 64 $\times$ 64   & 128      & 64 $\times$ 64        & 128                              \\ \hline
			Decoder1 & 128 $\times$ 128 & 64       & \multicolumn{2}{c}{\cellcolor{gray!50}}                 \\ \hline
		\end{tabu}%
	}
	\vspace{1mm}
	\caption{Encoder-decoder architecture of Exp2VRealFace.
		Layers in gray can be optionally removed to process $128 \times 128$ frames in real time by the entire EgoFace pipeline (Sec.~\ref{sec:results}).}
	\label{table:arch-cgan}
\end{table}

\section{Results}
\label{sec:results}

\begin{figure*}
	\includegraphics[width=\linewidth]{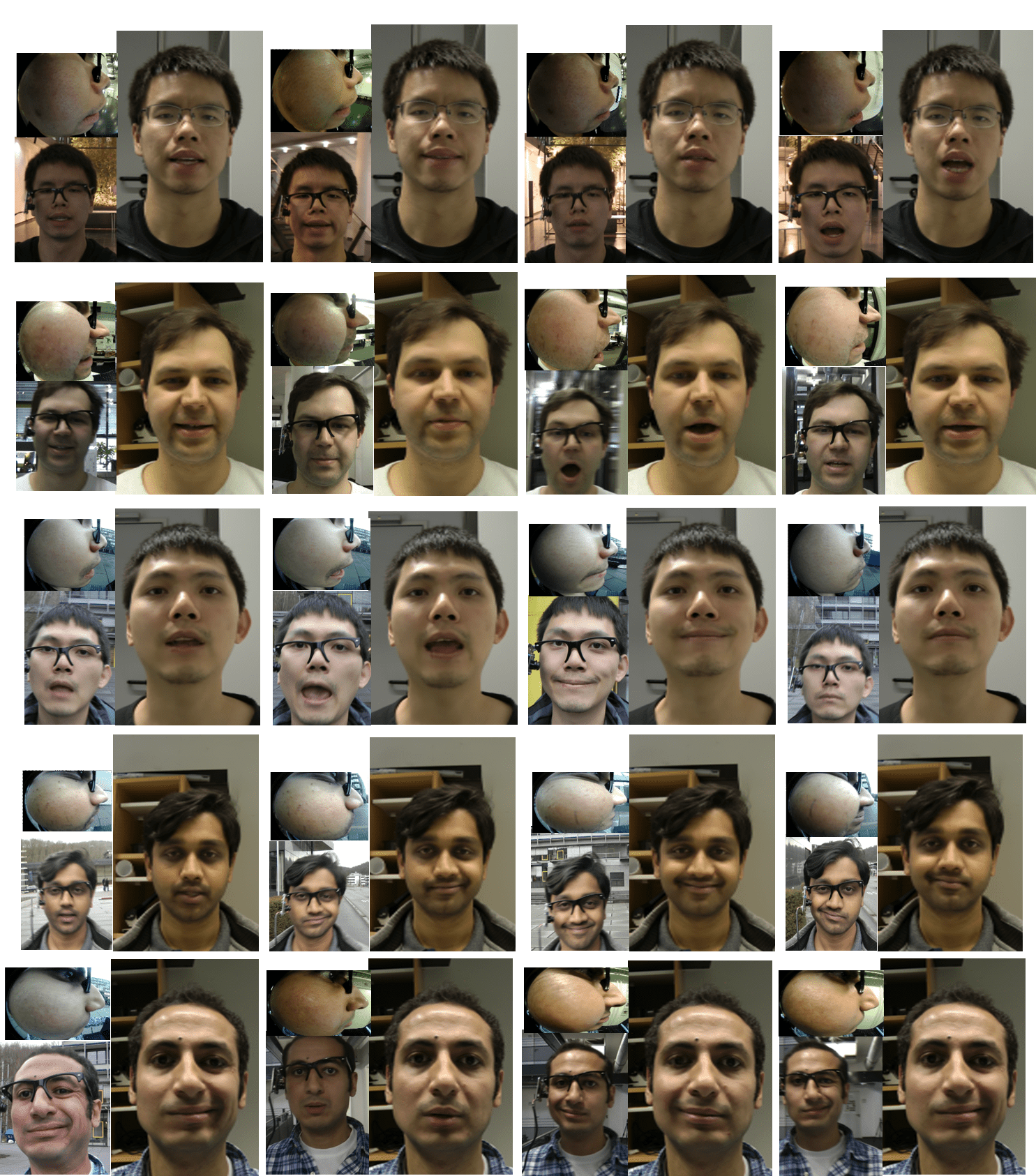}
	\caption{%
		Results for different people, shot under different illuminations, backgrounds for mobile indoor and outdoor environments. For each sample we show the egocentric view, which is the only input to EgoFace (left-top). The corresponding front view (left-bottom) shows the target expression. The final videorealistic animation produced by EgoFace is shown on the right. EgoFace captures a wide variety of expressions despite the highly oblique egocentric input. Please refer to the accompanying video for full results. 
	}
	\label{fig:TestAll}
\end{figure*}

\begin{figure*}
\centering
	\includegraphics[width=1\linewidth]{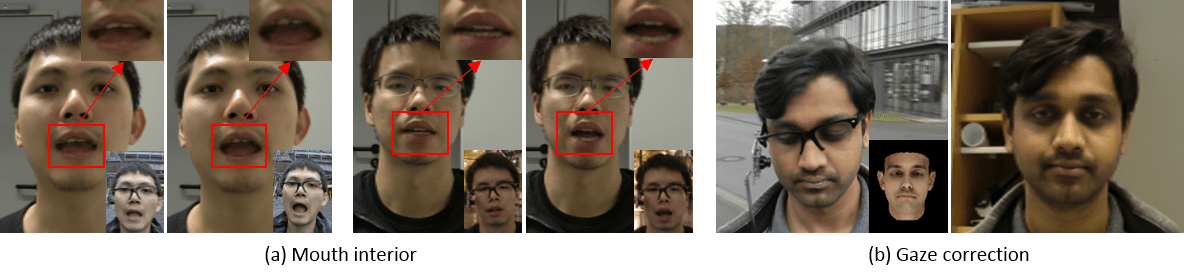}
	\caption{%
		(a) Consecutive frames produced by EgoFace. Our approach produces temporally consistent mouth interior, capturing both teeth and tongue. (b) The parametric face model allows gaze correction to better represent attention during a mobile teleconferencing application.
	}
	\label{fig:Details}
\end{figure*}

We evaluated Ego2Face on 5 people of different ethnicities moving in several indoor and outdoor environments. In total, we processed 24,500 frames shot in different environments. The facial expressions estimated by Ego2Exp are retargeted to four static scenes of different illuminations and backgrounds. Fig.~\ref{fig:TestAll} shows samples from sequences processed with our technique. For each frame, we show the input egocentric image, the corresponding front view and the videorealistic retargeting produced by Exp2VRealFace. Our approach produces temporally coherent videorealistic animation. Fig.~\ref{fig:TestAll} shows our approach handles a wide variety of facial expressions, including smiling, mouth opening, talking and others. Our approach handles asymmetrical expressions hardly visible from the egocentric view. This is because face expression are never entirely one-sided. Instead, a movement in one half of the face is usually correlated with the other half of the face. Our technique synthesizes a dynamic skin texture, by capturing the illumination as the skin deforms. Fig.~\ref{fig:TestAll} (last row) shows examples of this where our approach synthesizes skin texture triggered by stretching the face cheeks. For full results please see the accompanying video. 

Fig.~\ref{fig:Details} (a) shows examples where our approach handles the mouth interior in a temporally coherent manner. Here, both teeth and tongue are synthesized in a temporally consistent manner, without any special handling. Furthermore, the parametric face model give us full control over the human pose in the final animation. This allows us to correct the gaze for mobile teleconferencing applications. In such applications the user head is in continuous motion to watch out for the road. The parametric face model, however, enables us to correct the pose to better represent attention as shown in Fig.~\ref{fig:Details} (b).   

\textbf{Evaluation of face performance capture} To the best of our knowledge, current face reconstruction methods cannot deal with the highly distorted and occluded egocentric faces we use. 
State-of-the-art learning-based methods \cite{tewari2018pami, Tran2018b} train on large frontal image datasets. 
Because of the lack of such large-scale data for egocentric datasets, these methods cannot be easily extended to our use case.  
State-of-the-art optimization-based methods \cite{GarriZCVVPT2016, ThiesZSTN2016a} also assume a full perspective camera. 
These methods typically fail under large poses and self-occlusions, and thus are not suitable for our egocentric images.
In addition, most learning-based and optimization-based methods rely on image landmarks. 
Since there are no landmark detectors available for egocentric faces, current methods cannot be used in an automatic manner. 

\begin{figure}
\centering
	\includegraphics[width=1\linewidth]{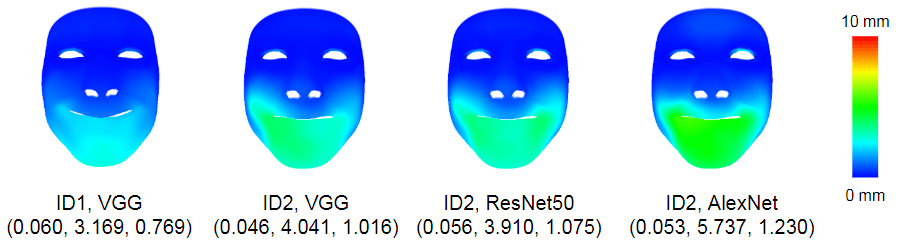}
	\caption{The average per-vertex Euclidean distance between Ego2Exp geometry and the ground-truth estimates 
	\cite{GarriZCVVPT2016}. For each image we show the (min., max., mean) values in mm averaged over 499 frames. While the ground-truth processes the front view, we achieve comparable results from our highly oblique egocentric view
	}
	\label{fig:RT3D}
\end{figure}

Our capture setup allows us to use existing reconstruction techniques to estimate the face parameters only from the corresponding front view. We use this as ground-truth for training our Ego2Exp (Sec.~\ref{sec:DataCollection}). We evaluate the results of Ego2Exp by comparing it to the estimates obtained using \cite{GarriZCVVPT2016} (Fig.~\ref{fig:RT3D}). 
We use the average per-vertex Euclidean distance between meshes.    
Note that while we use the highly distorted egocentric views as input, \cite{GarriZCVVPT2016} uses frontal views as input. 
Even then, we obtain comparable results demonstrating the high-quality expression reconstructions obtained by our approach.

\begin{figure}
\centering
	\includegraphics[width=\linewidth]{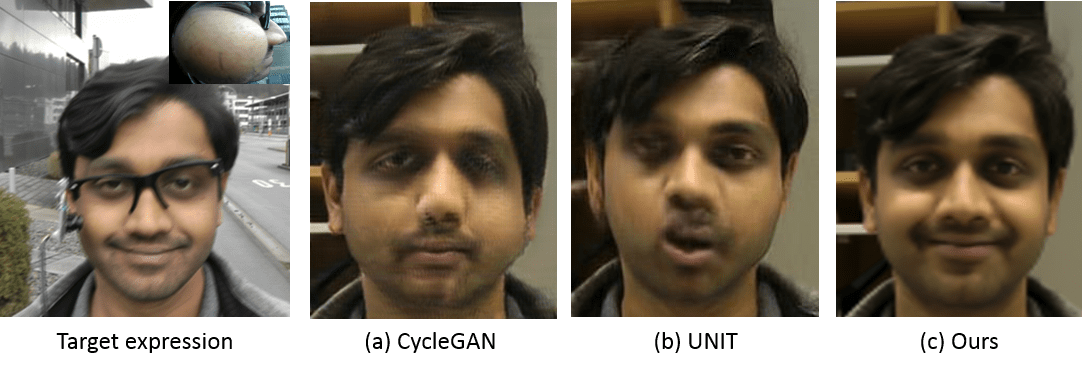}
	\caption{%
		Unpaired image-to-image translation techniques such as CycleGAN \cite{CycleGAN2017} and UNIT \cite{liu2017unsupervised} deform the face in an unnatural way. Our approach captures the target expression correctly and maintains the facial integrity. 
	}
	\label{fig:CycleGAN}
\end{figure}

\textbf{Comparing Retargeting to Related Work} To the best of our knowledge, there is no technique designed for handling our novel egocentric input. Face reenactment methods such as Face2Face and Deep Video Portraits \cite{ThiesZSTN2016a,kim18} rely on monocular face reconstruction approaches \cite{GarriZCVVPT2016,tewari2018pami} which do not work from our highly oblique egocentric view as discussed earlier. EgoFace does unpaired image translation. Even though Ego2Exp and Exp2VRealFace are each trained on paired data, the egocentric input to Ego2Exp and the front view output of Exp2VRealFace are not paired. We compare against publicly available implementations of unpaired image translation techniques such as CycleGAN \cite{CycleGAN2017} and UNIT \cite{liu2017unsupervised}. We trained these techniques on our egocentric and final front view face data. To remove the impact of the background, we took a tight crop around the face in both data. Fig.~\ref{fig:CycleGAN} shows that CycleGAN and UNIT deform the face in a very unnatural manner. Our approach, however, transfers the expressions correctly while maintaining the facial integrity. 

\begin{table}[]
	\centering
	\footnotesize
	\resizebox{0.35\textwidth}{!}{%
		\begin{tabu}{c|c|[1pt]c}
			\multicolumn{2}{c|[1pt]}{Components}        & Time (ms) \\ \tabucline[1pt]{-}
			\multirow{3}{*}{Ego2Exp}      & VGG         & 26.4       \\ \cline{2-3}
			                              & ResNet50      & 11.5      \\ \cline{2-3}
			                              & AlexNet     & 5.5       \\ \tabucline[1pt]{-}
			Synthetic rendering           & Albedo only & 3.3       \\ \tabucline[1pt]{-}
			\multirow{2}{*}{Exp2VRealFace} & Full       & 39.4       \\ \cline{2-3}
			                              & Optimized   & 21.4      \\ \hline
		\end{tabu}%
	}
	\vspace{1mm}
	\caption{Processing times of the different components of EgoFace.}
	\label{table:architecture}
\end{table}

\begin{figure}
	\centering
	\includegraphics[width=0.9\linewidth]{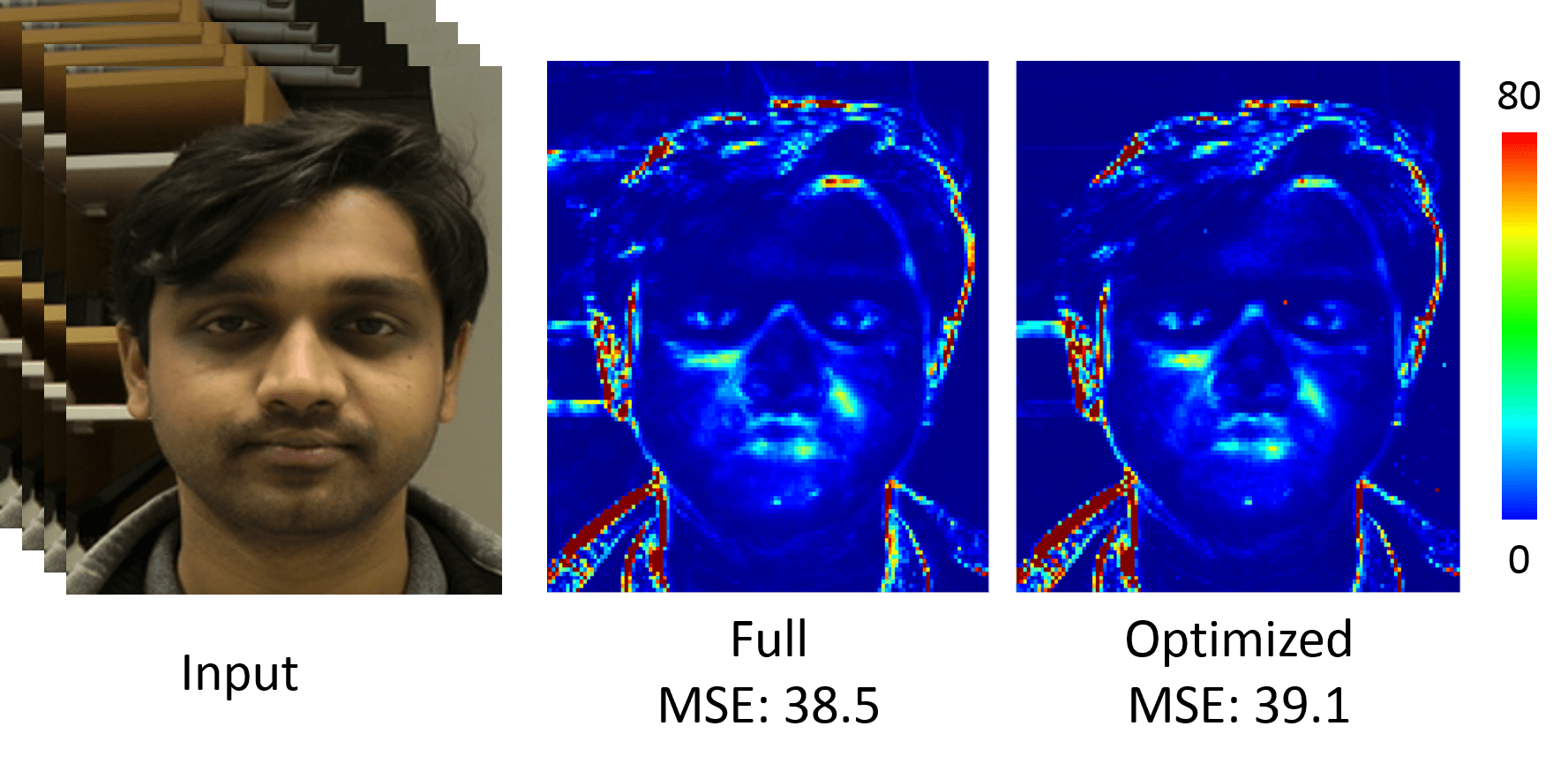}
	\caption{Per-pel mean square error between self reenactment and the input (ground-truth). We examine the full and optimized versions of Exp2VRealFace networks, with the same synthetic renderings. Results do not show significant difference between both networks.}
	\label{fig:AblExp2VRealFace}
\end{figure}

\textbf{Real-time implementation} As EgoFace lends itself to telepresence applications, we investigated running the entire pipeline in real time. We examined the computational efficiency of both Ego2Exp and Exp2VRealFace, with different implementations and network architectures. For Ego2Exp, we experimented with faster networks such as AlexNet and ResNet50. For Exp2VRealFace, we took a tight crop around the face and reduced the input image resolution to $128 \!\times\! 128$. This resolution is good to be viewed on a mobile phone screen, assuming a telepresence application. We removed some network layers and processed the input with no temporal window. Tab.~\ref{table:arch-cgan} shows the architecture of the optimized Syn2RealFace network. Tab.~\ref{table:architecture} shows the timings of each component of our EgoFace pipeline. We processed a sequence containing 4,500 images, twice, and took the average of their processing time. All network computations are performed on an NVIDIA Tesla V100 GPU. The synthetic rendering is performed on a Pascal Titan X. Our entire EgoFace pipeline can run in real-time at a rate of 36.2 ms per frame with ResNet50 and the optimized Exp2VRealFace. Fig.~\ref{fig:RT3D} shows ResNet50 achieves similar performance to VGG. Replacing ResNet50 with the less accurate AlexNet allows an even faster rate of 30.2 ms per frame, however, at the risk of slight temporal inconsistencies. Please refer to the accompanying video.

We compared the full and optimized Exp2VRealFace architectures numerically against ground-truth in a self-reenactment experiment. Given a sequence of 6,900 frames, we trained Exp2VRealFace on 5,000 and tested on the rest. We used the ground-truth albedo rendering as input for Exp2VRealFace. For fair numerical comparison we resized the output of the full network to $128 \!\times\! 128$. Both networks are trained with 200 epochs. Fig.~\ref{fig:AblExp2VRealFace} shows the mean squared error averaged over the 1,900 test frames. The average error for both architectures has no significant difference. 

\section{Conclusion}

We presented a new approach for face performance capture and reenactment. Current techniques use setups that limit applications in mobile environments. Our setup utilizes a lightweight single egocentric camera attached to a eye glass frame. This allows operations in dynamic and uncontrolled environments. Given a single, highly distorted one sided face view as input, we produce a videorealistic face animation of the user from the front view. Our approach handles people from different ethnicities, variable illuminations, backgrounds and a variety of expressions. It outperforms unpaired image translation techniques. Our setup is suitable for mobile video conferencing applications and can run in real time. Future work can address obtaining a single generic expression regression model. 

{\small
\bibliographystyle{ieee}
\bibliography{egbib}
}

\end{document}